\documentclass[10pt,twocolumn,letterpaper]{article}
\usepackage[pagenumbers]{cvpr}
\usepackage{tikz}
\usetikzlibrary{arrows.meta,positioning,calc}
\usepackage{array}
\usepackage{stfloats}

\definecolor{cvprblue}{rgb}{0.21,0.49,0.74}
\usepackage[pagebackref,breaklinks,colorlinks,allcolors=cvprblue]{hyperref}
\def\confName{CVPR}
\def\confYear{2027}
\title{Trusting Imagined Futures: Auditing Imagined Rewards \\for VLA Post-Training}
\author{
Zicheng Kong\textsuperscript{1},
Zhenbo Xu\textsuperscript{1,3},
Haoyan Zhang\textsuperscript{1},
Jinghan Yang\textsuperscript{1},
Zhenzhen Li\textsuperscript{1}\\
Yiwei Ru\textsuperscript{2},
Liuyu Xiang\textsuperscript{1},
WanTing Xu\textsuperscript{4},
Min Ren\textsuperscript{1},
Zhaofeng He\textsuperscript{1}\\[4pt]
{\small
\textsuperscript{1}Beijing University of Posts and Telecommunications,
Beijing, P.R. China}\\
{\small
\textsuperscript{2}Institute of Automation, Chinese Academy of Sciences,
Beijing, P.R. China}\\
{\small
\textsuperscript{3}ShiFang Technology Inc.,
Hangzhou, P.R. China}\quad
{\small
\textsuperscript{4}Wuhan Electric Power Technical College,
Wuhan, P.R. China}
}
\begin{document}
\maketitle

\begin{abstract}
Video world models enable candidate actions to be evaluated through imagined outcomes, but high predicted rewards can favor trajectories with poor execution quality.
We introduce Trusting Imagined Futures (TIF), an offline supervision selection method for vision-language-action (VLA) post-training that audits predicted rewards using execution feedback.
A multidimensional reward model predicts a utility score for candidate ranking and five dimensions of execution quality.
An auditor uses reward predictions and uncertainty to identify overestimation patterns learned from paired imagined videos and real executions.
TIF accepts the candidate with the highest predicted utility when its advantage over the existing trajectory exceeds a margin and its audit risk passes a threshold; otherwise, it retains the existing supervision.
The retained real trajectories then supervise behavior cloning.
Across four LIBERO suites and three training seeds, TIF achieves a mean success rate of 76.25\%, improving over training on the existing trajectories by 3.00 percentage points with the same trajectory count and training budget.
Adding audit gating yields 1.47 percentage points with the reward model and advantage margin held fixed.
The gains persist against advantage ranking and logistic regression risk selection at matched replacement counts.
On recorded videos from three real robot tasks, TIF increases clean completion among selected trajectories from 58.33\% to 70.00\% across 60 candidate pools.
\end{abstract}

\section{Introduction}
\label{sec:intro}

Vision-language-action (VLA) models generate robot actions from visual observations and language instructions, providing pretrained policies that can be adapted to different tasks~\cite{zitkovich2023rt2,octo2024,kim2024openvla,black2024pi0}.
When post-training relies on imitation learning, the quality of the supervision trajectories directly affects the behaviors the policy learns.
Available trajectories vary in task progress and execution quality, and imitating them can cause the policy to reproduce undesirable behaviors.
Selecting suitable supervision trajectories is therefore important for VLA post-training.

\begin{figure}[t]
    \centering
    \includegraphics[width=\columnwidth]{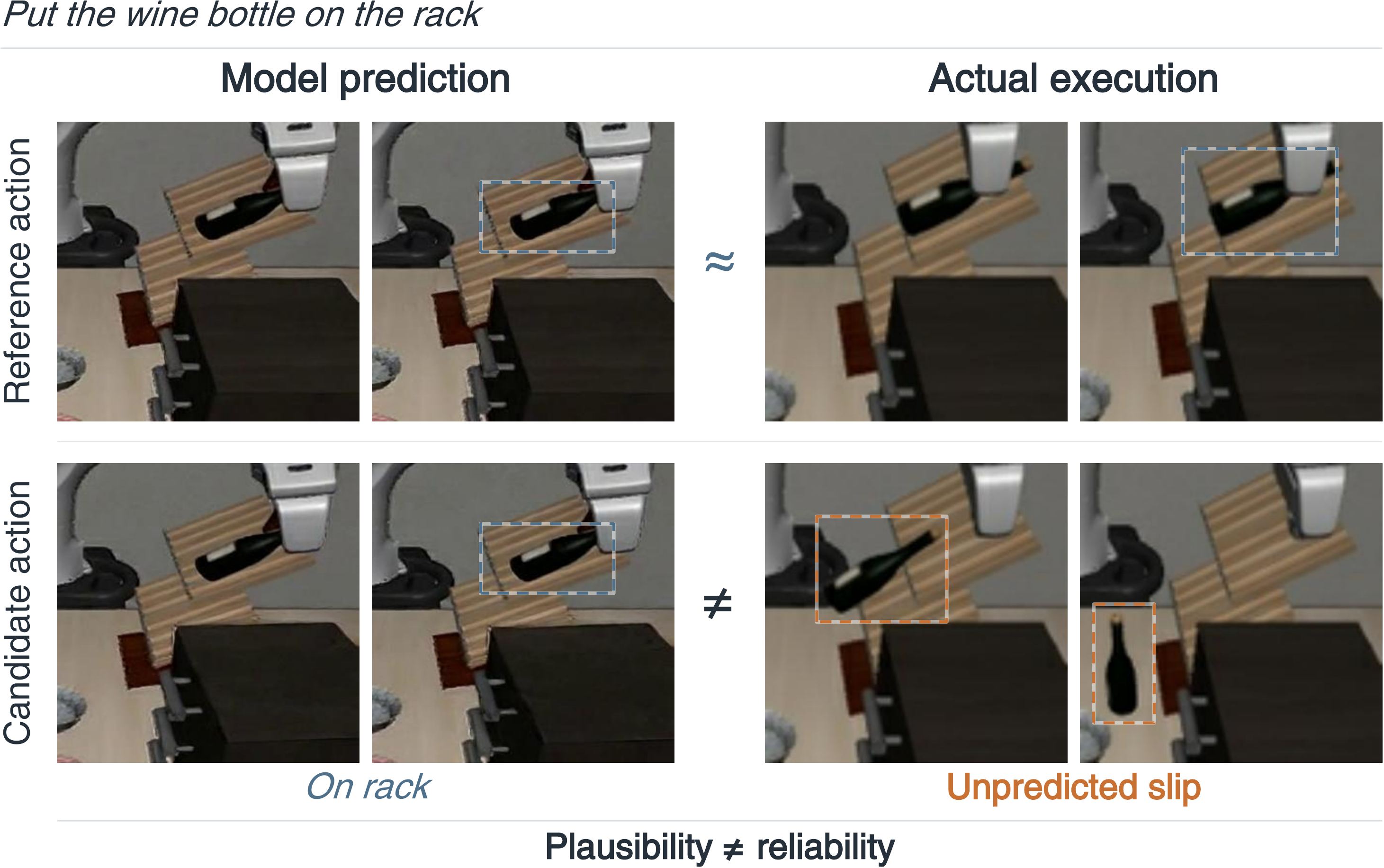}
    \caption{
        A candidate may receive a higher predicted utility than
        Source yet exhibit poorer execution quality.
        TIF audits evaluation reliability before accepting a replacement.
    }
    \label{fig:motivation}
\end{figure}

Video world models predict the visual consequences of candidate actions, providing imagined outcomes for behavior evaluation~\cite{yang2023unisim,zhu2025wmpo}.
A reward model scores these outcomes to compare candidates and support policy learning~\cite{zhu2025wmpo,zhang2026wise}.
When imagined outcomes are used to select supervision for VLA post-training, reward ranking identifies candidates with higher predicted utility.
Accepting such a candidate as a replacement also requires assessing whether its predicted advantage over the existing trajectory is trustworthy.

An apparently successful grasp or task completion in an imagined video may correspond to an unstable grasp, a collision, or an incorrect object interaction during real execution.
Errors in world model predictions and reward estimates can give such candidates high scores~\cite{gao2022overoptimization,jiang2026wovr}.
When they replace existing trajectories as behavior cloning targets, these evaluation errors can cause the policy to learn undesirable behaviors.
We study whether auditing rewards predicted from imagined outcomes improves supervision selection and subsequent policy learning (Fig.~\ref{fig:motivation}).
We evaluate this question using fixed candidate pools with matched executions.

We propose Trusting Imagined Futures (TIF), which uses real execution feedback to train a risk auditor that complements reward scores when deciding whether to accept candidate trajectories as supervision.
The reward model evaluates imagined videos along five dimensions: task progress, goal correctness, manipulation quality, safety, and execution stability.
It jointly predicts a utility score for candidate ranking.
The auditor learns from discrepancies between these predictions and the corresponding real execution outcomes to identify candidates with high predicted scores but poor execution quality.
TIF first selects the candidate with the highest predicted utility.
It accepts this candidate if its predicted advantage over the existing supervision trajectory (Source) exceeds a margin and its audit risk falls below a threshold; otherwise, it retains Source.
We then update the VLA through behavior cloning using paired observations and actions from the retained real execution trajectories.

We construct 32,000 pairs of real executions and imagined videos across four LIBERO suites~\cite{liu2023libero}, using real execution outcomes to supervise the reward model and auditor.
Candidate pools are collected offline, with recorded executions providing labels, evaluation outcomes, and behavior cloning targets.
Selection uses predicted scores and diagnostic information from imagined videos to accept a replacement or retain Source.
At deployment, only the updated VLA is required.

Across four LIBERO suites and three training seeds, TIF achieves a mean success rate of 76.25\%, exceeding behavior cloning using only Source trajectories by 3.00 percentage points with the same number of training trajectories and behavior cloning budget.
With the reward model and advantage margin held fixed, adding audit gating improves mean success by 1.47 percentage points.
When the number of replacements is fixed, TIF still outperforms ranking by predicted advantage and selecting candidates by risk estimated through logistic regression.

Our main contributions are as follows:
\begin{itemize}
    \item We propose TIF, a supervision auditing method that learns from matched real executions when rewards predicted from imagined videos overestimate execution quality. TIF combines this risk with a candidate's predicted advantage to decide whether to accept a replacement or retain the existing supervision.

    \item We develop a multidimensional reward model for candidate ranking and risk auditing. The model jointly predicts five dimensions of execution quality and a utility score. Utility predictions rank candidates, while dimension predictions and their uncertainty provide information for auditing.

    \item We show that the auditor identifies discrepancies between high predicted rewards and poor execution outcomes. Comparisons with fixed replacement counts and downstream post-training experiments further show that using audit risk improves supervision selection and policy success.
\end{itemize}

\section{Related Work}

\paragraph{VLA Post-Training and World Models.}
OpenVLA-OFT improves supervised adaptation through changes to action decoding, representation, and learning objectives~\cite{kim2025openvlaoft}, while GRAPE aligns policies with preferences over successful and failed trajectories~\cite{zhang2024grape}.
RIPT-VLA and VLA-RL use environment interaction for reinforcement learning, with sparse success rewards and process rewards, respectively~\cite{tan2025riptvla,lu2025vlarl}.
WMPO reduces reliance on real environment interaction by optimizing policies through imagined rollouts from a video world model~\cite{zhu2025wmpo}.
WoVR addresses accumulated prediction errors through keyframe initialization and joint updates of the policy and world model~\cite{jiang2026wovr}.
WoVR+ extends this approach to multiple views, proprioceptive states, and training within individual stages of longer tasks~\cite{zhou2026wovrplus}.
TIF uses imagined outcomes to evaluate fixed candidate actions and select real execution trajectories for behavior cloning.

\paragraph{Robot Reward Models and Candidate Evaluation.}
Robometer learns rewards through progress supervision within trajectories and preference comparisons across trajectories~\cite{liang2026robometer}, while RoboReward augments supervision with negative and partial completion examples constructed through counterfactual relabeling and temporal clipping~\cite{lee2026roboreward}.
RynnValue learns temporal distance to task completion and converts its predictions into dense rewards~\cite{huang2026rynnvalue}.
During inference, V-GPS reranks actions using a value function learned through offline reinforcement learning~\cite{nakamoto2024vgps}, and RoVer uses process rewards to guide candidate expansion and selection~\cite{dai2025rover}.
RARM gates progress rewards according to confidence in matching execution clips to a reference demonstration~\cite{yang2026rarm}.
TIF uses matched simulator executions to supervise predictions of execution quality from imagined videos, providing a utility score for ranking and five dimension predictions for auditing.

\paragraph{Reliability of Imagined Supervision.}
Recent methods use reliability estimates to control how imagined outcomes influence policy learning.
WISE filters candidate groups using reward differences, ranking consistency, and reward variation across repeated imaginations, then updates the policy using relative advantages within each group~\cite{zhang2026wise}.
HaWMPO learns video reliability from visual, geometric, and motion discrepancies against observed executions, together with image quality metrics, and adjusts reinforcement learning rewards through Reward-Soft~\cite{chen2026hawmpo}.
MeRL uses visual prediction and reward proxy errors to regulate imagination length and data proportion, as well as sampling probabilities and training weights for individual chunks~\cite{li2026merl}.
TIF learns the risk of reward overestimation from matched executions, targeting cases with high predicted utility, low executed utility, and a substantial gap between them.
This risk and the predicted advantage over Source determine whether to accept the candidate with the highest predicted utility or retain the existing supervision.

\begin{figure*}[!t]
    \centering
    \includegraphics[width=0.95\textwidth]{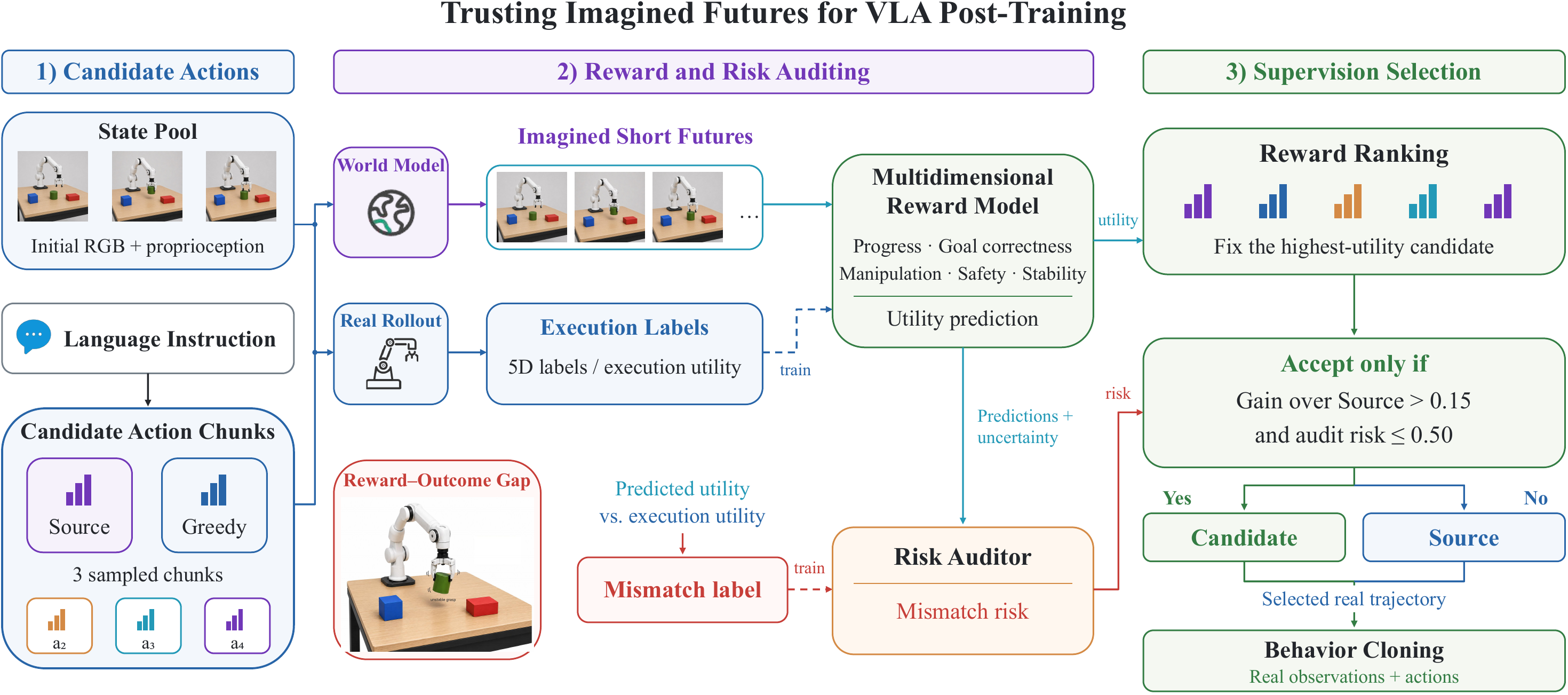}
    \caption{
    Overview of TIF.
    Matched executions supervise reward learning and reliability
    auditing.
    During curation, predicted utility and audit risk determine
    whether to replace Source.
    The selected real trajectories supervise VLA post-training
    across all valid steps.
    }
    \label{fig:overview}
\end{figure*}

\section{Method}
\label{sec:method}

\paragraph{Overview.}
As shown in Fig.~\ref{fig:overview}, TIF combines candidate utility prediction with reward risk auditing to select supervision trajectories for VLA post-training.
We first train a multidimensional reward model using matched execution outcomes, then train an auditor on discrepancies between its predictions and those outcomes.
During selection, the candidate with the highest predicted utility replaces Source only if its predicted advantage exceeds a margin and its audit risk falls below a threshold; otherwise, Source is retained.
Observations and actions from all valid steps of the retained real execution trajectories are then used for behavior cloning.

\subsection{Candidate Actions and Paired Outcomes}
\label{sec:paired_futures}

Given a visual observation $o_i$, proprioceptive state $q_i$, and task instruction $\ell_i$ at initial state $i$, we construct a candidate action set $\mathcal{A}_i=\{\mathbf{a}_{i,k}\}_{k=0}^{K-1}$.
Each sequence contains at most eight control commands executed at 20 Hz.
The component dataset contains four controlled candidates per initial state, providing varied execution quality for training and evaluation.
The natural pool used for supervision selection contains five candidates: Source, one greedy branch, and three sampled branches.

Source is a fixed action proposal associated with the initial state, taken from subsequent actions in a LIBERO demonstration or from saved policy action records.
Sequences are extended when necessary according to the rules
described in Appendix~\ref{app:source}.
We execute Source in its associated scene and initial state.
The remaining branches use the OpenVLA policy for the corresponding suite to generate actions step by step from the current execution observation, using greedy decoding or sampling.
For every branch, we record observations, actions, and execution feedback until the sequence reaches its prescribed length or the environment terminates.

We collect candidates offline and keep their actions and execution records fixed when comparing selection rules.
The world model generates an imagined video $\hat{\tau}_{i,k}$ from the initial observation and recorded action sequence, with an adapter converting initial proprioception into the required conditioning.
For candidates that terminate early, we use only the valid executed action prefix.
The matched execution trajectory $\tau_{i,k}$ provides training labels, outcomes for evaluating selection decisions, and paired observations and actions for behavior cloning.
Selection uses predicted scores and diagnostic information from imagined videos.

\subsection{Reward Supervision from Execution}
\label{sec:reward_supervision}

Final task success does not fully capture local progress and execution quality within short action sequences.
We therefore derive five continuous labels from each candidate's execution:
$\mathbf{r}_{i,k}=(P_{i,k},G_{i,k},M_{i,k},S_{i,k},E_{i,k})\in[0,1]^5$,
representing task progress, goal correctness, manipulation quality, a safety proxy, and execution stability.

Task progress measures advancement toward completion, while goal correctness measures alignment with the object and goal specified by the instruction.
Manipulation quality measures the effectiveness of grasping, transport, and placement.
The safety proxy captures risks associated with abnormal contact, excessive actions, and abrupt state changes.
Execution stability measures motion smoothness, action reversals, and gripper switching.
These labels are computed from execution feedback, states, contacts, and action statistics, with normalization parameters fitted on the training set.
The continuous labels are used to construct reference utility and are discretized into low, medium, and high levels using thresholds of 0.34 and 0.67 for training the dimension predictions.
Dimension definitions and discretization are summarized
in Appendix~\ref{app:dimension_labels}.

We construct reference utility using predefined weights for task progress, goal correctness, manipulation quality, and execution stability, together with a multiplicative safety factor:
\begin{equation}
\begin{aligned}
v_{i,k}
={}&
\bigl(
0.35P_{i,k}
+0.30G_{i,k}
+0.20M_{i,k}
+0.15E_{i,k}
\bigr) \\
&\cdot
\bigl(0.35+0.65S_{i,k}\bigr).
\end{aligned}
\label{eq:reference_utility}
\end{equation}
We cap $v_{i,k}$ at 0.666 when neither task progress nor goal correctness reaches the high level, and at 0.339 when the safety label is low.
The lower cap applies when both conditions hold, yielding the final reference utility $u_{i,k}$.
This utility supervises the utility prediction head and defines preferences between candidates from the same initial state.

\subsection{Multidimensional Reward Learning and Risk Auditing}
\label{sec:reward_auditing}

We use the dimension labels and reference utility defined in Sec.~\ref{sec:reward_supervision} to train a reward model that predicts execution quality from imagined videos, task instructions, and candidate actions.
We then train a risk auditor using discrepancies between predicted utility and the reference utility from matched executions.
After reward training, its parameters remain fixed throughout auditor training and trajectory selection.

\paragraph{Multidimensional reward model.}
The reward model $R_\theta$ takes the imagined video $\hat{\tau}_{i,k}$, task instruction $\ell_i$, and candidate action sequence $\mathbf{a}_{i,k}$ as input.
A DINOv2-L encoder with fixed parameters extracts video frame features, which are fused with instruction and action information.
The model predicts distributions over the low, medium, and high levels of each reward dimension, along with a utility estimate $\hat{u}_{i,k}$ from a separate prediction head.
The utility estimate ranks candidates, while the dimension predictions provide information for risk auditing.

We jointly optimize dimension prediction, utility regression, and pairwise ranking:
\begin{equation}
\mathcal{L}_R
=
\mathcal{L}_{\mathrm{dim}}
+
0.5\mathcal{L}_{\mathrm{utility}}
+
0.1\mathcal{L}_{\mathrm{pair}}.
\label{eq:reward_loss}
\end{equation}
The three losses supervise the predicted dimension levels, agreement with execution reference utility, and the relative ordering of candidates from the same initial state, respectively.

\paragraph{Auditing reward overestimation.}
After training the reward model, we compute its predictions on a separate dataset for auditor training.
We assign a positive audit label to candidates with high predicted utility, low execution reference utility, and a sufficiently large overestimation gap:
\begin{equation}
y_{i,k}
=
\mathbf{1}\!\left[
\begin{aligned}
&\hat{u}_{i,k}\geq 0.58
\;\land\;
u_{i,k}\leq 0.42\\
&\land\;
\hat{u}_{i,k}-u_{i,k}\geq 0.20
\end{aligned}
\right].
\label{eq:audit_target}
\end{equation}

The auditor $C_\psi$ takes a feature vector $\mathbf{z}_{i,k}$ containing predicted utility, distributions and scores for the five dimensions, uncertainty statistics, and basic diagnostics from imagined videos.
These diagnostics describe video availability, image quality,
and changes between frames; the video diagnostic fields are
listed in Appendix~\ref{app:auditor_features}.
The auditor outputs a risk score $\rho_{i,k}=C_\psi(\mathbf{z}_{i,k})$ and is trained with $y_{i,k}$ as its binary target.
Matched execution outcomes provide the training labels, while reward predictions and imagined video diagnostics provide the inputs for estimating overestimation risk.

\subsection{Reliability Gating and Policy Learning}
\label{sec:curation_posttraining}

TIF first selects the candidate with the highest predicted utility, then decides whether to accept it based on its predicted advantage over Source and its audit risk.
For initial state $i$, we assign index $0$ to Source and define the candidate with the highest predicted utility and its advantage over Source:
\begin{equation}
k_i^*=\arg\max_k \hat{u}_{i,k},
\qquad
\Delta_i=\hat{u}_{i,k_i^*}-\hat{u}_{i,0}.
\label{eq:candidate_winner}
\end{equation}
We accept the candidate when its predicted advantage exceeds $0.15$ and its audit risk is at most $0.50$; otherwise, we retain Source.
The retained candidate is
\begin{equation}
\tilde{k}_i=
\begin{cases}
k_i^*, & \Delta_i>0.15 \;\land\; \rho_{i,k_i^*}\leq0.50,\\
0, & \text{otherwise}.
\end{cases}
\label{eq:selection_gate}
\end{equation}
Each initial state contributes one training trajectory, provided by either the accepted candidate or Source.

\paragraph{Behavior cloning.}
We train the policy using the real execution trajectory corresponding to the selected candidate $\tilde{k}_i$.
Its valid observations and actions are denoted by $\mathcal{T}_i=\{(o_{i,t},a_{i,t})\}_{t=0}^{L_i-1}$, where $o_{i,t}$ is the observation before action $a_{i,t}$ and $L_i$ is the number of valid steps.
During training, we sample a trajectory uniformly and then sample one of its valid time steps uniformly.
For a selected dataset containing $N$ trajectories, the behavior cloning objective is
\begin{equation}
\mathcal{L}_{\mathrm{BC}}
=
-\mathbb{E}_{\substack{
i\sim\mathcal{U}(\{1,\ldots,N\})\\
t\sim\mathcal{U}(\{0,\ldots,L_i-1\})
}}
\left[
\log\pi_\phi(a_{i,t}\mid o_{i,t},\ell_i)
\right].
\label{eq:segment_bc}
\end{equation}
For OpenVLA, we compute the loss over the corresponding action tokens and update only the LoRA parameters in the attention and MLP modules.
At deployment, the updated VLA directly generates actions from observations and task instructions.

\section{Experiments}
\label{sec:experiments}

\subsection{Experimental Setup}
\label{sec:experimental_setup}

\paragraph{Tasks and models.}
We evaluate TIF on 40 tasks across four LIBERO suites: Spatial, Object, Goal, and Long~\cite{liu2023libero}.
We use OpenVLA as the policy~\cite{kim2024openvla}, Cosmos-Predict2 2B to generate imagined videos~\cite{nvidia2025cosmos}, and DINOv2-L to extract visual features for the reward model~\cite{oquab2023dinov2}.

\paragraph{Data.}
The dataset for reward model and auditor training and evaluation contains 32,000 pairs of execution trajectories and imagined videos from 8,000 initial states.
The natural candidate dataset supports offline selection evaluation and policy training, with 1,600 initial states split into 800/400/400 for training, validation, and testing.
Offline evaluation uses the 384 test states in which the task is not already successful at initialization.
Data splits are grouped by source trajectory, with reward model and auditor training data separated from natural validation and test data by trajectory provenance.

\paragraph{Baselines.}
We compare against OpenVLA before post-training and policies trained by behavior cloning on Source, greedy branches, or randomly selected candidates.
Reward-only selects the candidate with the highest predicted utility.
Reward+Margin additionally requires its predicted advantage over Source to exceed the margin, and TIF further applies the audit risk gate.
At matched replacement counts, we also compare random selection, predicted advantage ranking, logistic regression risk, prediction entropy, and consistency scores.
All post-training comparisons use the same candidate pool, policy initialization, and training budget, retaining one real execution trajectory per initial state.

\paragraph{Training and evaluation.}
We adapt the policy through LoRA~\cite{hu2021lora} and behavior cloning using 200 selected training trajectories per suite, for 800 trajectories in total.
Training uses all valid time steps by default; supervision from only the first action is evaluated as an ablation.
Each suite uses 500 updates and an effective batch size of 16 observation--action pairs.
We train each method with three seeds and evaluate each resulting policy over 50 trials per task, totaling 2,000 episodes per seed.
Evaluation initial states and environment seeds are identical across methods.
We report policy success as the mean and sample standard deviation across the three training seeds.
Data splits, policy training settings, and seed-wise results
for matched replacement comparisons are provided in
Appendices~\ref{app:data_splits},
\ref{app:selection_training}, and \ref{app:policy_seeds},
respectively.

\subsection{Multidimensional Reward Evaluation}
\label{sec:reward_evaluation}

We first examine whether the reward model predicts actual execution quality from imagined videos and whether dimensional supervision improves candidate ranking.
We evaluate 1,920 candidates from 384 test states against labels and reference utilities derived from matched executions.
Macro-F1 ranges from 0.732 to 0.758 across the five dimensions; results for individual dimensions and ordinal errors are provided in Appendix~\ref{app:dimension_results}.

We compare Scalar, which predicts utility alone, with 2D, which adds supervision for task progress and goal correctness, and 5D, which uses all five dimensions.
The three base models share the same reference utility and pairwise supervision.
Each model selects candidates by predicted utility without an advantage margin or audit gate.
As shown in \cref{tab:reward_evaluation}, 5D achieves the highest pairwise accuracy, Top-1 accuracy, and mean execution reference utility of selected candidates, at 57.79\%, 33.07\%, and 0.4603, respectively.
We further ablate pairwise supervision and separately shuffle the manipulation, safety, and stability labels.
Shuffling these labels reduces Top-1 accuracy to 29.17\% and selected utility to 0.4550, indicating that their correspondence with execution outcomes contributes to candidate ranking.

We further examine the information provided by imagined videos.
Using the same candidate set, we separately train reward models with the initial image, instruction, and actions, either alone or supplemented with imagined or recorded execution videos.
Candidates are selected by predicted utility without a margin or audit gate.
Adding imagined videos improves Top-1 accuracy from 29.95\% to 33.07\% and the mean execution utility of selected candidates from 0.4527 to 0.4603, showing that predicted visual changes provide additional information for candidate evaluation.
Recorded execution videos yield a higher Top-1 accuracy of 50.52\%, showing the gap between evaluation from predicted and observed outcomes.
Full input comparisons are provided in Appendix~\ref{app:input_information}.
We next evaluate whether auditing reward overestimation improves supervision selection based on imagined videos.

\begin{table}[!tb]
    \centering
    \caption{
        \textbf{Candidate ranking with different reward supervision.}
        Pairwise and Top-1 accuracies are reported in \%.
        Top-1 measures whether the selected candidate has maximal
        execution reference utility within its candidate set.
        Utility is the mean execution reference utility of selected candidates.
        The final two rows independently ablate 5D.
    }
    \label{tab:reward_evaluation}
    \small
    \setlength{\tabcolsep}{3pt}
    \begin{tabular*}{\columnwidth}{
        @{\extracolsep{\fill}}lccc@{}
    }
        \toprule
        Model
        & Pairwise $\uparrow$
        & Top-1 $\uparrow$
        & Utility $\uparrow$ \\
        \midrule
        Scalar
        & 56.88 & 30.99 & .4574 \\
        2D
        & 57.45 & 32.55 & .4598 \\
        5D
        & \textbf{57.79} & \textbf{33.07} & \textbf{.4603} \\
        \midrule
        5D w/o pairwise
        & 57.42 & 30.99 & .4598 \\
        5D shuffled M/S/E
        & 56.80 & 29.17 & .4550 \\
        \bottomrule
    \end{tabular*}
\end{table}

\subsection{Reliability Gating}
\label{sec:reliability_evaluation}

The candidate with the highest predicted reward can still have lower execution utility than the existing supervision.
Using the same 5D reward model on 384 test states, Reward-only makes 310 replacements, of which 219 have lower execution reference utility than Source.
With the advantage margin and audit gate, TIF reduces the fraction of replacements that lower utility from 70.6\% to 53.2\%, while maintaining a similar mean utility (0.4602 versus 0.4603 for Reward-only).
\Cref{tab:reliability_evaluation}(a) reports replacement outcomes for each rule.
Mean utility includes states that retain Source.

Audit risk distinguishes candidates with high predicted utility but poor execution outcomes among the reward winners.
Of the 310 non-Source candidates selected by reward ranking, 23 satisfy the mismatch definition in \cref{eq:audit_target}.
This event captures substantial reward overestimation, whereas the Worse counts above include any replacement with execution utility below Source.
The auditor identifies these events with an AUROC of 0.77.
Mismatch rates are 22.58\% in the highest risk quintile and 1.61\% in the lowest.

We further compare selection rules at the same replacement count to evaluate the contribution of risk ranking.
Each rule selects 111 of the same 310 candidates, without applying an advantage margin.
As shown in \cref{tab:reliability_evaluation}(b), audit risk selects 58 candidates with higher execution reference utility than Source, compared with 47 for predicted advantage and 52 for logistic regression risk, and achieves the highest mean utility.
Event diagnostics and extended selection comparisons are
provided in Appendices~\ref{app:risk_identification}
and \ref{app:fixed_replacements}, respectively.

\begin{table}[!tb]
    \centering
    \caption{
        \textbf{Supervision selection with reliability auditing.}
        Better/Worse count replacements with higher/lower execution
        reference utility than Source.
        Utility averages the retained trajectories over all 384 states.
        (a) Default selection rules.
        (b) Each rule accepts 111 of the same 310 non-Source reward
        winners without an advantage margin and retains Source elsewhere.
        Random reports expected values.
    }
    \label{tab:reliability_evaluation}
    \small
    \setlength{\tabcolsep}{3pt}

    \begin{tabular*}{\columnwidth}{
        @{\extracolsep{\fill}}lcccc@{}
    }
        \toprule
        \multicolumn{5}{@{}l}{
            \textbf{(a) Default selection rules}
        } \\
        \midrule
        Method
        & Replaced
        & Better
        & Worse
        & Utility $\uparrow$ \\
        \midrule
        Source
        & 0 & 0 & 0 & .4444 \\
        Reward-only
        & 310 & 91 & 219 & \textbf{.4603} \\
        Reward+Margin
        & 186 & 65 & 121 & .4587 \\
        TIF
        & 111 & 52 & 59 & .4602 \\
        \bottomrule
    \end{tabular*}

    \smallskip

    \begin{tabular*}{\columnwidth}{
        @{\extracolsep{\fill}}lccc@{}
    }
        \toprule
        \multicolumn{4}{@{}l}{
            \textbf{(b) Fixed replacement count: 111}
        } \\
        \midrule
        Selection criterion
        & Better $\uparrow$
        & Worse $\downarrow$
        & Utility $\uparrow$ \\
        \midrule
        Random (expected)
        & 32.58 & 78.42 & .4501 \\
        Predicted advantage
        & 47 & 64 & .4589 \\
        Logistic regression risk
        & 52 & 59 & .4621 \\
        Auditor risk
        & \textbf{58} & \textbf{53} & \textbf{.4636} \\
        \bottomrule
    \end{tabular*}
\end{table}


\begin{table*}[!t]
    \centering
    \caption{
        \textbf{Policy success on LIBERO (\%).}
        Post-training methods share the same policy initialization,
        trajectory count, and training budget, using all valid time steps.
        Values are means $\pm$ sample standard deviations across three
        training seeds.
        Original OpenVLA is evaluated before post-training.
        Bold indicates the highest mean in each column.
    }
    \label{tab:policy_results}
    \small
    \setlength{\tabcolsep}{4pt}
    \begin{tabular*}{\textwidth}{
        @{\extracolsep{\fill}}lccccc@{}
    }
        \toprule
        Method & Spatial & Object & Goal & Long & Average \\
        \midrule
        Original OpenVLA
        & 74.80
        & 86.20
        & 77.20
        & 37.80
        & 69.00 \\
        \midrule
        Source
        & $80.20 \pm 0.92$
        & $86.07 \pm 1.03$
        & $\mathbf{83.73 \pm 0.31}$
        & $43.00 \pm 1.06$
        & $73.25 \pm 0.68$ \\
        Greedy
        & $78.00 \pm 1.11$
        & $84.53 \pm 0.76$
        & $81.67 \pm 0.42$
        & $41.93 \pm 1.33$
        & $71.53 \pm 0.51$ \\
        Random
        & $79.07 \pm 1.21$
        & $85.33 \pm 0.42$
        & $82.67 \pm 0.42$
        & $41.93 \pm 1.21$
        & $72.25 \pm 0.48$ \\
        \midrule
        Reward-only
        & $81.60 \pm 0.72$
        & $87.40 \pm 0.72$
        & $82.93 \pm 0.64$
        & $45.00 \pm 1.20$
        & $74.23 \pm 0.40$ \\
        Reward+Margin
        & $82.13 \pm 1.01$
        & $88.27 \pm 0.50$
        & $83.33 \pm 0.50$
        & $45.40 \pm 1.39$
        & $74.78 \pm 0.63$ \\
        TIF
        & $\mathbf{84.13 \pm 0.83}$
        & $\mathbf{89.40 \pm 0.72}$
        & $82.73 \pm 0.64$
        & $\mathbf{48.73 \pm 0.83}$
        & $\mathbf{76.25 \pm 0.58}$ \\
        \bottomrule
    \end{tabular*}
\end{table*}

\begin{table*}[!t]
    \centering
    \caption{
        \textbf{Audit targets and supervision ablations.}
        Policy success is averaged over four suites
        (\%, mean $\pm$ sample SD across three training seeds).
        (a) Auditors share inputs, training data, and training budget.
        Offline selection accepts 111 of 310 test proposals without
        a margin; Worse counts replacements below Source, and Utility
        averages all 384 test states including Source fallback.
        Policy training separately uses 230 replacements from training
        candidates that pass the margin, with counts matched by suite.
        (b) Each reward model has its own auditor trained under the
        same protocol.
        (c) Selected trajectories and training budgets are fixed
        within each selection method.
    }
    \label{tab:policy_ablations}
    \small
    \setlength{\tabcolsep}{4pt}

    \begin{tabular*}{\textwidth}{
        @{\extracolsep{\fill}}lccc@{}
    }
        \toprule
        \multicolumn{4}{@{}l}{
            \textbf{(a) Audit training target}
        } \\
        \midrule
        Target
        & \multicolumn{2}{c}{Offline selection}
        & Policy success $\uparrow$ \\
        \cmidrule(lr){2-3}
        & Worse $\downarrow$ & Utility $\uparrow$ & \\
        \midrule
        Signed error regression
        & 56
        & .4624
        & $75.60 \pm 0.46$ \\
        Below-Source classification
        & \textbf{50}
        & .4629
        & $75.85 \pm 0.59$ \\
        High-score mismatch classification
        & 53
        & \textbf{.4636}
        & $\mathbf{76.25 \pm 0.58}$ \\
        \bottomrule
    \end{tabular*}

    \medskip

    \begin{minipage}[t]{0.40\textwidth}
        \centering
        \begin{tabular*}{\linewidth}{
            @{\extracolsep{\fill}}lc@{}
        }
            \toprule
            \multicolumn{2}{@{}l}{
                \textbf{(b) Reward supervision}
            } \\
            \midrule
            Supervision & Policy success $\uparrow$ \\
            \midrule
            Scalar & $74.57 \pm 0.33$ \\
            2D & $75.43 \pm 0.48$ \\
            5D & $\mathbf{76.25 \pm 0.58}$ \\
            \bottomrule
        \end{tabular*}
    \end{minipage}
    \hfill
    \begin{minipage}[t]{0.56\textwidth}
        \centering
        \begin{tabular*}{\linewidth}{
            @{\extracolsep{\fill}}lcc@{}
        }
            \toprule
            \multicolumn{3}{@{}l}{
                \textbf{(c) Trajectory supervision}
            } \\
            \midrule
            Selection & First action & All valid steps \\
            \midrule
            Source
            & $71.97 \pm 0.80$
            & $73.25 \pm 0.68$ \\
            TIF
            & $74.18 \pm 0.61$
            & $\mathbf{76.25 \pm 0.58}$ \\
            \bottomrule
        \end{tabular*}
    \end{minipage}
\end{table*}

\subsection{Policy Improvement through Supervision Selection}
\label{sec:policy_improvement}

We next evaluate whether the trajectories selected in Sec.~\ref{sec:reliability_evaluation} improve policy learning, using task success after behavior cloning.
With the same trajectory count and behavior cloning budget, TIF achieves an average success rate of 76.25\%, exceeding Source by 3.00 percentage points (\cref{tab:policy_results}).
Training directly on greedy branches or random candidates does not outperform Source, whereas reward selection improves average success.
With the reward model and advantage margin held fixed, adding the audit gate yields a further gain of 1.47 percentage points, with improvements across all three training seeds.
TIF achieves the highest success on Spatial, Object, and Long.
Its largest gain over Source is 5.73 percentage points on Long, while success on Goal decreases by 1.00 percentage point.

Section~\ref{sec:reliability_evaluation} evaluates risk ranking at a fixed acceptance count; we extend this comparison to policy training.
All selection rules use the same training candidates that pass the advantage margin and match TIF's replacement count within each suite, yielding 230 replacement trajectories and 570 Source trajectories.
TIF outperforms predicted advantage ranking at 75.55\% and logistic regression risk at 75.65\%, as well as random, low entropy, and consistency selection (\cref{tab:matched_policy_baselines}).
Its success rate is higher than each control across all three training seeds.

\begin{table}[!t]
    \centering
    \caption{
        \textbf{Policy success at matched replacement counts.}
        All rules select from the same 348 training candidates that
        pass the advantage margin, accepting 57/61/54/58 replacements
        in Spatial/Object/Goal/Long.
        Each training set contains 230 replacements and 570 Source
        trajectories.
        Selections remain unchanged across training seeds.
        Success is averaged over all four suites
        (\%, mean $\pm$ sample SD).
        TIF uses the result from \cref{tab:policy_results}.
    }
    \label{tab:matched_policy_baselines}
    \small
    \renewcommand{\arraystretch}{1.1}
    \setlength{\tabcolsep}{4pt}
    \begin{tabular*}{\columnwidth}{
        @{\extracolsep{\fill}}lc@{}
    }
        \toprule
        Selection rule & Success $\uparrow$ \\
        \midrule
        Random-K
        & $74.15 \pm 0.63$ \\
        Advantage Top-K
        & $75.55 \pm 0.50$ \\
        Logistic risk-K
        & $75.65 \pm 0.52$ \\
        Low-entropy-K
        & $75.03 \pm 0.51$ \\
        Consistency-K
        & $75.60 \pm 0.52$ \\
        TIF
        & $\mathbf{76.25 \pm 0.58}$ \\
        \bottomrule
    \end{tabular*}
\end{table}

\paragraph{Audit targets and supervision.}
To examine the role of the mismatch target, we compare it with signed error regression and classification of candidates whose execution utility is below Source.
The auditors share the same inputs, training data, and training budget.
Each target is evaluated through the offline selection protocol in Sec.~\ref{sec:reliability_evaluation} and the policy training comparison at matched replacement counts.
As shown in \cref{tab:policy_ablations}(a), classification relative to Source retains the fewest replacements that lower execution utility, while mismatch supervision achieves the highest mean execution utility.
Mismatch supervision also yields the highest policy success, at 76.25\%, compared with 75.85\% for classification relative to Source and 75.60\% for signed error regression.

Section~\ref{sec:reward_evaluation} shows that dimensional supervision improves candidate evaluation.
Within the complete selection and policy training pipeline, average success increases from 74.57\% with Scalar to 75.43\% with 2D and 76.25\% with 5D (\cref{tab:policy_ablations}b).
These variants share the same reference utility, gating rules, and policy training budget, with a separate auditor trained for each reward model.
Detailed results on the Long suite are provided
in Appendix~\ref{app:policy_long}.

With selected trajectories and training budgets fixed, using all valid steps instead of only the first action improves success by 1.28 percentage points for Source and 2.07 for TIF (\cref{tab:policy_ablations}c).
TIF outperforms Source with either scheme.



\begin{figure*}[!t]
    \centering
    \includegraphics[width=0.9\textwidth]
    {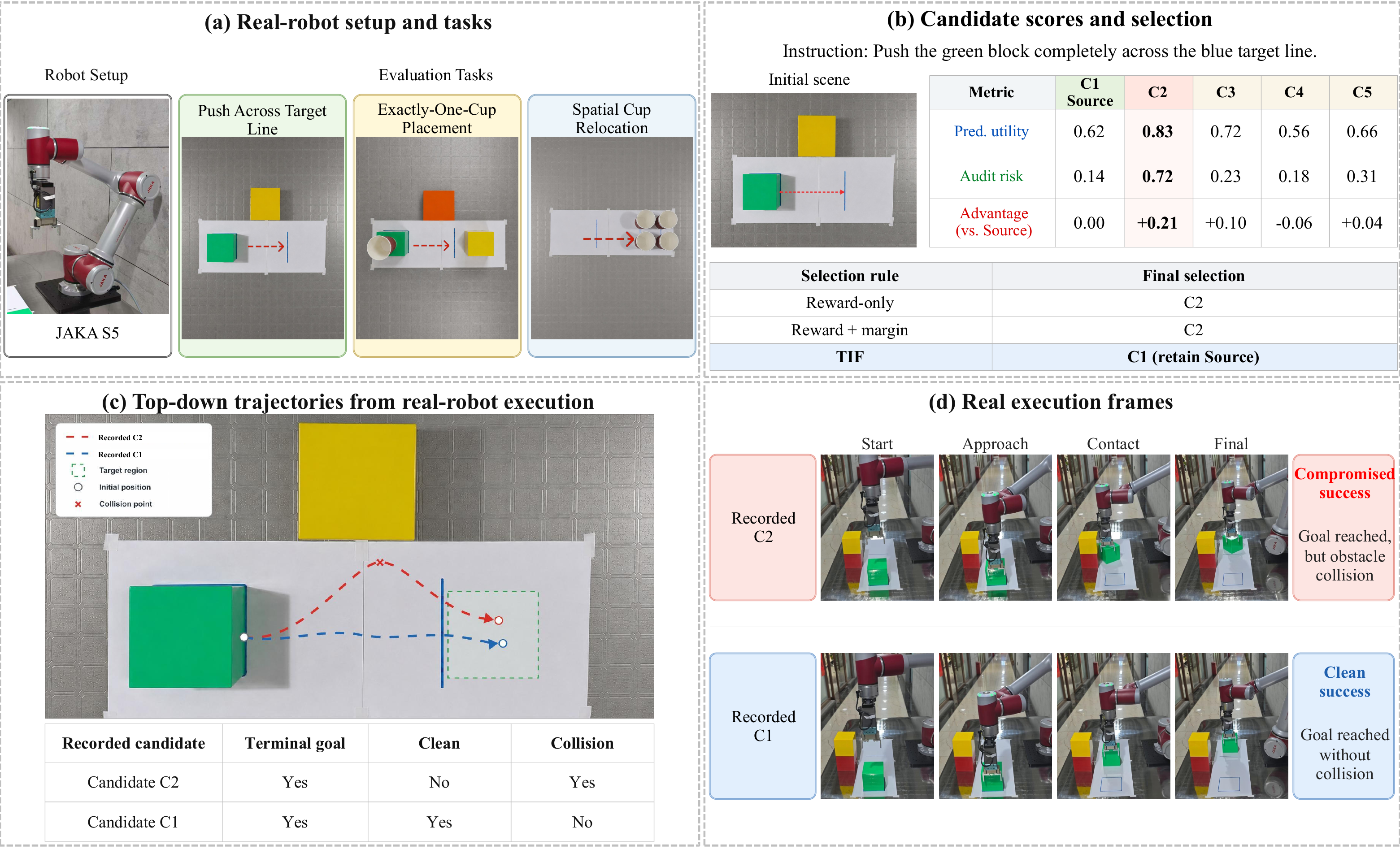}
    \caption{
Real-robot candidate selection.
(a) Tasks and setup. (b) Scores and selections, with advantage relative to Source C1: TIF rejects C2 and retains C1.
(c,d) Recorded executions: both reach the goal, but only C2 collides with the obstacle.
    }
    \label{fig:real_robot}
\end{figure*}

\subsection{Supervision Selection on a Real Robot}
\label{sec:real_robot}

We evaluate trajectory selection on a JAKA S5 robot using three tasks: pushing a block across a target line, placing exactly one cup, and relocating a specified cup (\cref{fig:real_robot}a).
Each task has 20 candidate pools with five physical executions per pool, including Source, yielding 300 trajectories.
Both methods score recorded execution videos with the reward model used above; TIF applies the same auditor and acceptance rule.
Task definitions and the evaluation protocol are provided
in Appendices~\ref{app:robot_tasks}
and \ref{app:robot_selection}.

TIF increases task completion among selected trajectories from 75.00\% to 78.33\% and clean completion from 58.33\% to 70.00\% (\cref{tab:real_robot_results}).
Clean completion satisfies the instruction with a stable final state and no collision, wrong or extra object interaction, or unnecessary recovery.

In the pushing example, C2 exceeds the advantage margin but fails the audit gate, so TIF retains Source C1 (\cref{fig:real_robot}b).
Both complete the task, but C2 collides with the obstacle while C1 completes the push without collision (\cref{fig:real_robot}c,d).

\begin{table}[!t]
    \centering
    \caption{
        \textbf{Selection outcomes on recorded robot videos.}
        Counts refer to selected trajectories, with one selection
        per candidate pool.
        Reward denotes Reward-only.
    }
    \label{tab:real_robot_results}
    \small
    \setlength{\tabcolsep}{3pt}
    \begin{tabular*}{\columnwidth}{
        @{\extracolsep{\fill}}lcccc@{}
    }
        \toprule
        & \multicolumn{2}{c}{Task completion}
        & \multicolumn{2}{c}{Clean completion} \\
        \cmidrule(lr){2-3}
        \cmidrule(lr){4-5}
        Task & Reward & TIF & Reward & TIF \\
        \midrule
        Push
        & 15/20 & 16/20 & 11/20 & 14/20 \\
        One cup
        & 14/20 & 15/20 & 11/20 & 13/20 \\
        Relocation
        & 16/20 & 16/20 & 13/20 & 15/20 \\
        \midrule
        Total
        & 45/60 & 47/60 & 35/60 & 42/60 \\
        \bottomrule
    \end{tabular*}
\end{table}

\section{Conclusion}
\label{sec:conclusion}

We presented TIF, which combines multidimensional reward evaluation with risk auditing learned from execution feedback to select supervision trajectories for VLA post-training.
The reward model provides utility predictions for ranking and dimension predictions with uncertainty for auditing.
The retained real trajectories supervise behavior cloning.
With the same trajectory count and training budget, TIF improves average success across four LIBERO suites from 73.25\% to 76.25\%.
Dimensional supervision ablations and comparisons at fixed replacement counts support the contributions of execution quality information and risk auditing to policy learning.

\paragraph{Limitations.}
Experiments use OpenVLA and LIBERO.
Future work will examine other policies and benchmarks, and the execution feedback needed to adapt the auditor when models or candidate distributions change.

\clearpage

{
    \small
    \bibliographystyle{ieeenat_fullname}
    \bibliography{main}
}

\clearpage
\setcounter{page}{1}
\maketitlesupplementary
\appendix

\setcounter{figure}{0}
\setcounter{table}{0}
\renewcommand{\thefigure}{S\arabic{figure}}
\renewcommand{\thetable}{S\arabic{table}}

\section{Data and Candidate Construction}
\label{app:data}

\subsection{Tasks and Data Splits}
\label{app:data_splits}

Our experiments cover the Spatial, Object, Goal, and Long suites of LIBERO, with ten tasks per suite~\cite{liu2023libero}.
We construct a component dataset for training and evaluating the reward model and auditor, and a natural candidate dataset for supervision selection and policy post-training.
In both datasets, candidates associated with the same initial state form a candidate pool.
Each candidate has a recorded execution trajectory and a corresponding imagined video.
Table~\ref{tab:supp_data} summarizes the dataset sizes and their roles.

\begin{table*}[t]
    \centering
    \caption{Dataset sizes and roles. Candidate counts are per initial state. Each pair consists of a recorded execution trajectory and its corresponding imagined video.}
    \label{tab:supp_data}
    \small
    \setlength{\tabcolsep}{5pt}
    \begin{tabular*}{\textwidth}{@{\extracolsep{\fill}}lrrrp{0.35\textwidth}@{}}
        \toprule
        Dataset & Initial states & Candidates & Pairs & Role \\
        \midrule
        Component & 8,000 & 4 & 32,000 & Reward model and auditor training and evaluation \\
        Natural training & 800 & 5 & 4,000 & Supervision selection for behavior cloning \\
        Natural validation & 400 & 5 & 2,000 & Validation and configuration selection \\
        Natural testing & 400 & 5 & 2,000 & Offline candidate evaluation and selection analysis \\
        \bottomrule
    \end{tabular*}
\end{table*}

Data splits are grouped by source trajectory.
Initial states derived from the same source trajectory, together with all their candidates, are assigned to the same split.
When records have multiple levels of derivation, we trace their provenance back to the original trajectory.
The natural validation and test sets inherit the validation and test provenance assignments of the component dataset, respectively.
Neither set shares source trajectories with the training data of the reward model or auditor.
The natural training set may share source trajectories with the reward training data.

Offline reward evaluation and auditing analyses use the 384 natural test states in which the task is not already successful at initialization, yielding 1,920 candidates.
Policy post-training uses the natural training set and retains one execution trajectory per initial state, giving 800 trajectories in total and 200 per suite.

\subsection{Candidate Construction and Execution}
\label{app:source}
\label{app:candidates}

The component dataset contains four controlled candidates per initial state, providing examples with different execution quality.
Each natural candidate pool contains five candidates: Source, one greedy branch, and three sampled branches.

Source is an existing fixed action proposal associated with the initial state, obtained from either a LIBERO demonstration or saved policy action records.
For demonstration sources, we extract the action sequence following the corresponding starting point.
For historical policy sources, we retrieve the saved actions.
The construction of historical policy records includes extending a sequence by repeating its final action, as well as repeating a saved single action.
We then execute Source in its associated scene and initial state, recording observations, actions, and execution feedback.

The remaining four branches use the OpenVLA policy for the corresponding suite~\cite{kim2024openvla}.
The greedy branch uses greedy decoding, while the three sampled branches use temperature $1$, top-$p$ $1$, and top-$k$ $0$.
At each step, the policy generates an action from the current execution observation, allowing subsequent actions to respond to changes observed during execution.

All branches within a pool start from the same scene and initial state.
Each sequence contains at most eight control commands executed at 20\,Hz and stops upon reaching its prescribed length or environment termination.
For a branch that terminates early, we retain the valid executed prefix.
After collection, the action sequences and their execution records remain fixed for subsequent video generation, evaluation, and supervision selection.

\subsection{Pairing Imagined Videos with Executions}
\label{app:paired_outcomes}

We use Cosmos-Predict2 2B to generate an imagined video for each candidate~\cite{nvidia2025cosmos}.
Generation is conditioned on the initial visual observation and the recorded action sequence, with an adapter using the initial proprioceptive state to construct the required world model conditioning.
For candidates that terminate early, generation uses the valid action prefix.

Each pair associates an imagined video and an execution record with the same task instruction, initial state, and candidate action sequence.
The execution record contains observations, actions, and feedback at valid steps, which are used to construct the five quality labels and reference utility.
The label computations are described in Appendix~\ref{app:labels}.

Offline selection uses reward predictions from imagined videos and the corresponding audit risk.
One candidate is retained per initial state, and observation--action pairs from all valid steps of its actual execution segment are used for behavior cloning.
World model conditioning and video processing details are provided in Appendix~\ref{app:implementation}.

\section{Execution Labels and Evaluation Metrics}
\label{app:labels}

\subsection{Five Dimensions of Execution Quality}
\label{app:dimension_labels}

We construct five continuous quality scores from each candidate's
actual execution: task progress, goal correctness, manipulation
quality, safety, and execution stability.
The labels use environment feedback, robot and object states,
contact records, and action statistics, with normalization
parameters fitted on the training data.
Each score lies in $[0,1]$, with higher values indicating better
quality in the corresponding dimension.
Table~\ref{tab:supp_dimensions} summarizes their definitions
and evaluation signals.

\begin{table*}[t]
    \centering
    \caption{
        Execution quality dimensions and the signals used to
        construct their labels.
    }
    \label{tab:supp_dimensions}
    \small
    \setlength{\tabcolsep}{5pt}
    \begin{tabular}{@{}p{0.18\textwidth}
                      p{0.29\textwidth}
                      p{0.46\textwidth}@{}}
        \toprule
        Dimension & Evaluation focus & Execution signals \\
        \midrule
        Task progress ($P$)
        & Advancement toward task completion
        & Changes in success status, environment rewards,
          approach to the target, and relevant object state changes
        \\[3pt]

        Goal correctness ($G$)
        & Agreement with the object and goal specified by the instruction
        & Target object interaction, approach to the goal location,
          and satisfaction of task conditions
        \\[3pt]

        Manipulation quality ($M$)
        & Effectiveness of grasping, transport, and placement
        & Gripper engagement, object motion, end-effector motion,
          and feedback on task-relevant operations
        \\[3pt]

        Safety ($S$)
        & Risks associated with abnormal contact and motion
        & Contact anomalies, excessive actions, action saturation,
          and abrupt state changes
        \\[3pt]

        Execution stability ($E$)
        & Smoothness of actions and state changes
        & Action smoothness, action reversals, state jitter,
          and gripper switching
        \\
        \bottomrule
    \end{tabular}
\end{table*}

Task progress and goal correctness describe the extent of progress
and its alignment with the instruction, respectively.
Object motion may indicate local progress, while goal correctness
further considers whether the manipulated object, motion direction,
and resulting state match the instruction.
Manipulation quality evaluates the effectiveness of the interaction.
Safety and execution stability capture abnormal risks and the
smoothness of execution, respectively.

The continuous scores are used to construct reference utility,
while discrete levels supervise dimension prediction.
All five dimensions use the same thresholds:
scores below $0.34$ are low, scores from $0.34$ to below $0.67$
are medium, and scores of at least $0.67$ are high.
These levels are encoded as $0$, $1$, and $2$, respectively.

\subsection{Execution Reference Utility}
\label{app:reference_utility}

We combine the five continuous scores into an execution reference
utility to supervise utility prediction and compare candidates
from the same initial state.
We first compute
\begin{equation}
\begin{aligned}
v ={}&
\bigl(0.35P + 0.30G + 0.20M + 0.15E\bigr) \\
&\cdot \bigl(0.35 + 0.65S\bigr).
\end{aligned}
\label{eq:supp_reference_utility}
\end{equation}
Task progress, goal correctness, manipulation quality, and
execution stability contribute through a weighted sum.
Safety adjusts this sum through a multiplicative factor.

We then apply two caps.
When neither task progress nor goal correctness reaches the high
level, the score is capped at $0.666$.
When safety is low, the score is capped at $0.339$.
The lower cap applies when both conditions hold.
The resulting value is the final reference utility $u$.
Within each initial state, candidates with higher reference utility
receive higher preference.

The reward model outputs a utility estimate $\hat{u}$ through a
separate prediction head and uses this estimate to rank candidates.
The dimension predictions provide additional information for auditing.
Equation~\ref{eq:supp_reference_utility} constructs the supervision
target; predicted utility is produced directly by the utility head.

\subsection{Audit Targets and Evaluation Events}
\label{app:audit_targets}

After reward training, we compute predictions on a separate dataset
for auditor training and pair them with the execution reference
utilities.
A candidate is labeled as a high-score mismatch when all three
conditions hold:
\begin{equation}
\begin{aligned}
\hat{u} &\geq 0.58, \\
u &\leq 0.42, \\
\hat{u}-u &\geq 0.20.
\end{aligned}
\label{eq:supp_mismatch}
\end{equation}
These conditions require high predicted utility, low execution
utility, and a sufficiently large overestimation gap.
The audit label is $1$ when all conditions hold and $0$ otherwise.
The reward model parameters remain fixed during auditor training.

Additional analyses compare two alternative audit targets.
Signed error uses the difference $\hat{u}-u$ as a regression target.
Below Source uses a binary target indicating whether the candidate's
execution reference utility is lower than that of Source from the
same initial state.

High-score mismatch measures disagreement between prediction and
execution, whereas Below Source compares the replacement candidate
with the existing supervision.
The two events may occur together or separately.
Subsequent tables report their positive counts and discrimination
metrics separately.

\subsection{Evaluation Metrics}
\label{app:metrics}

We evaluate reward prediction, candidate selection, and policy
performance using the metrics in Table~\ref{tab:supp_metrics}.

\begin{table*}[t]
    \centering
    \caption{
        Evaluation metrics and their aggregation.
        Execution reference utility is used for candidate evaluation;
        task success is used for downstream policy evaluation.
    }
    \label{tab:supp_metrics}
    \small
    \setlength{\tabcolsep}{5pt}
    \begin{tabular}{@{}p{0.20\textwidth}p{0.74\textwidth}@{}}
        \toprule
        Metric & Definition \\
        \midrule

        Macro-F1
        & For each quality dimension, the unweighted mean of the
          F1 scores for the low, medium, and high levels.
        \\[3pt]

        Ordinal MAE
        & Mean absolute error between predicted and reference
          levels encoded as $0$, $1$, and $2$.
        \\[3pt]

        Pairwise accuracy
        & The fraction of candidate comparisons for which predicted
          utility agrees with the ordering given by execution
          reference utility. Comparisons are made within the
          same initial state.
        \\[3pt]

        Top-1 accuracy
        & The fraction of initial states for which the candidate
          selected by predicted utility has the highest execution
          reference utility in its pool.
        \\[3pt]

        Selected utility
        & Mean execution reference utility of the final retained
          candidates across initial states, including states
          that retain Source.
        \\[3pt]

        Better / Worse
        & Counts of replacements whose execution reference utility
          is higher or lower than that of Source, respectively.
        \\[3pt]

        AUROC / AP
        & Discrimination metrics for audit risk on the specified
          binary event, reported with the evaluation population
          and positive count.
        \\[3pt]

        Policy success rate
        & The fraction of successful episodes when evaluating
          the updated policy on complete tasks.
        \\
        \bottomrule
    \end{tabular}
\end{table*}

Top-1 accuracy measures utility ranking within candidate pools.
Policy success rate measures the ability of the updated policy
to complete tasks.
These metrics correspond to supervision selection and downstream
policy evaluation, respectively.

In the audit analyses, all proposals refers to candidates that
have the highest predicted utility in their pool and are not Source.
The high-score subset further requires predicted utility of at
least $0.58$.
AUROC and AP are computed using the event labels within the
corresponding population.

Policy results are computed separately for three training seeds
and reported as the mean success rate and sample standard deviation.
Each seed is evaluated over 2,000 episodes across the four suites,
with 500 episodes per suite.
Because the suites have equal evaluation sizes, the average success
rate across suites also equals the total success count divided
by 2,000 for that seed.

\section{Model and Training Details}
\label{app:implementation}

\subsection{Model Components}
\label{app:model_components}

TIF uses Cosmos-Predict2 2B to generate imagined videos,
DINOv2-L to extract visual features from video frames,
and OpenVLA as the policy for post-training
\cite{nvidia2025cosmos,oquab2023dinov2,kim2024openvla}.
The world model generates videos from the initial visual
observation, recorded action sequence, and conditioning
constructed from the initial proprioceptive state.
Candidate collection and pairing are described
in Appendix~\ref{app:data}.

The reward model fuses video features with the task instruction
and candidate action information.
It outputs a distribution over the low, medium, and high levels
for each of the five quality dimensions, together with a separate
utility estimate.
Utility predictions rank candidates, while dimension predictions
and associated uncertainty statistics support risk auditing.
The DINOv2-L parameters remain unchanged during reward training.

The risk auditor is a multilayer perceptron with approximately
25,000 trainable parameters.
It takes reward model outputs and imagined video diagnostics
as input and produces a mismatch risk score for each candidate.

\subsection{Reward Model and Auditor Training}
\label{app:reward_auditor_training}

The reward model is jointly trained using the dimension labels,
execution reference utility, and preferences between candidates
from the same initial state, as defined
in Appendix~\ref{app:labels}.
The training objective is
\begin{equation}
\mathcal{L}_{R}
=
\mathcal{L}_{\mathrm{dim}}
+
0.5\mathcal{L}_{\mathrm{utility}}
+
0.1\mathcal{L}_{\mathrm{pair}}.
\label{eq:supp_reward_loss}
\end{equation}
The dimension loss supervises the levels of the five quality
dimensions, the utility loss supervises overall utility prediction,
and the pairwise loss supervises the relative ordering of
candidates from the same initial state.

After reward training, we compute predictions on a separate
dataset for auditor training.
These predictions and their corresponding execution reference
utilities define the high-score mismatch labels
in Appendix~\ref{app:audit_targets}.
The auditor is trained using these binary labels.
The reward model parameters remain unchanged during this stage,
so the auditor learns the prediction errors of a fixed evaluator.

The trained reward model and auditor are then applied to the
natural candidate pools for evaluation and selection.
The complete Scalar, 2D, and 5D pipeline comparisons train a
corresponding reward model and auditor for each configuration,
using the same utility supervision, selection rules,
and policy training budget.

\subsection{Auditor Inputs and Video Diagnostics}
\label{app:auditor_features}

The auditor input has two components.
The first contains reward model outputs: predicted utility,
distributions and scores for the five dimensions,
and prediction uncertainty statistics.
The second contains imagined video diagnostics describing
availability, format, image quality, and changes between frames.
The selector separately computes the predicted advantage
over Source.

The video diagnostics contain 20 fields, grouped by function
in Table~\ref{tab:supp_video_diagnostics}.
All fields are extracted from imagined videos and use
the prefix \texttt{vq\_\_}.

\begin{table*}[t]
    \centering
    \caption{
        Imagined video diagnostic fields.
        The shared prefix \texttt{vq\_\_} is omitted.
    }
    \label{tab:supp_video_diagnostics}
    \small
    \setlength{\tabcolsep}{5pt}
    \begin{tabular}{@{}p{0.17\textwidth}
                      p{0.39\textwidth}
                      p{0.36\textwidth}@{}}
        \toprule
        Category & Fields & Description \\
        \midrule

        Availability and decoding
        & \texttt{feature\_available},
          \texttt{decode\_ok}
        & Feature availability and successful video decoding.
        \\[4pt]

        Video format
        & \texttt{width},
          \texttt{height},
          \texttt{fps},
          \texttt{frame\_count\_meta},
          \texttt{frame\_count\_decoded}
        & Width, height, frame rate, metadata frame count,
          and actual decoded frame count.
        \\[4pt]

        Brightness and contrast
        & \texttt{mean\_brightness},
          \texttt{mean\_contrast}
        & Image brightness and contrast statistics.
        \\[4pt]

        Sharpness
        & \texttt{mean\_laplacian\_var},
          \texttt{median\_laplacian\_var}
        & Mean and median Laplacian variance.
        \\[4pt]

        Temporal changes
        & \texttt{mean\_frame\_absdiff},
          \texttt{max\_frame\_absdiff},
          \texttt{first\_last\_absdiff}
        & Mean and maximum absolute differences between adjacent
          frames, and the absolute difference between the first
          and last frames.
        \\[4pt]

        Diagnostic flags
        & \texttt{is\_low\_motion},
          \texttt{is\_very\_low\_motion},
          \texttt{is\_low\_contrast},
          \texttt{is\_dark\_or\_blank},
          \texttt{is\_blurry\_proxy},
          \texttt{is\_bad\_decode}
        & Low motion, very low motion, low contrast,
          dark or blank images, a blur proxy, and decoding failure.
        \\
        \bottomrule
    \end{tabular}
\end{table*}

These diagnostics describe video readability, image quality,
and temporal changes, providing auxiliary information for auditing
reward predictions.
Supervision for execution quality, including collision risks
and goal correctness, comes from the paired execution records.

Candidate source metadata is excluded from the auditor input.
Execution reference utility is used to construct training labels
and evaluate selection outcomes.
Offline selection uses reward predictions, video diagnostics,
and audit risk.

\subsection{Supervision Selection and Policy Training}
\label{app:selection_training}

For each natural candidate pool, the selector first identifies
the candidate with the highest predicted utility and computes
its predicted advantage over Source.
The candidate is accepted when its advantage exceeds $0.15$
and its audit risk is at most $0.50$.
Otherwise, Source is retained.
If the candidate with the highest predicted utility fails
these conditions, the selector falls back directly to Source.
Each initial state contributes one actual execution segment.

Behavior cloning uses observation--action pairs from all valid
steps of the retained segments.
Each training sample is drawn by first selecting a segment
uniformly and then selecting a valid time step uniformly
within that segment.
The real observation before the action is used to predict
the corresponding action.
This procedure gives each segment the same probability
of being sampled.

We compute the training loss over OpenVLA action tokens and
update only the LoRA parameters in its attention and MLP modules.
All methods use the same policy initialization, candidate pools,
training trajectory count, and update budget.
For each method, the selected trajectory set remains unchanged
across the three training seeds.
Table~\ref{tab:supp_policy_training} summarizes the configuration.

\begin{table*}[t]
    \centering
    \caption{
        Policy post-training and evaluation configuration.
    }
    \label{tab:supp_policy_training}
    \small
    \setlength{\tabcolsep}{5pt}
    \begin{tabular}{@{}p{0.27\textwidth}p{0.67\textwidth}@{}}
        \toprule
        Item & Setting \\
        \midrule

        Policy
        & OpenVLA
        \\[3pt]

        Updated parameters
        & LoRA parameters in attention and MLP modules
        \\[3pt]

        Training data
        & One retained actual execution segment per initial state
          in the natural training set
        \\[3pt]

        Training segments
        & 200 per suite; 800 in total
        \\[3pt]

        Supervision steps
        & All valid execution steps
        \\[3pt]

        Sampling
        & Uniform over segments, then uniform over valid
          time steps within the selected segment
        \\[3pt]

        Updates
        & 500 per suite and training seed
        \\[3pt]

        Effective batch size
        & 16 observation--action pairs
        \\[3pt]

        Training seeds
        & 7, 11, and 23
        \\[3pt]

        Evaluation episodes
        & 50 per task; 500 per suite
        \\[3pt]

        Evaluation initial conditions
        & The same evaluation initial states and environment
          seeds across methods
        \\
        \bottomrule
    \end{tabular}
\end{table*}

During complete-task evaluation, the updated OpenVLA policy
generates actions independently.
The reward model, world model, and auditor are used for offline
supervision selection and do not participate in action generation
during this evaluation.

\section{Additional Reward Evaluation}
\label{app:reward}

This section supplements the reward evaluation in
Sec.~\ref{sec:reward_evaluation}.
All experiments use 1,920 candidates from the 384 natural test
states in which the task is not already successful at initialization.
Quality labels and reference utilities are derived from the
paired executions.
Selection uses the candidate with the highest predicted utility,
without an advantage margin or audit gate.

\subsection{Prediction of Individual Dimensions}
\label{app:dimension_results}

Table~\ref{tab:supp_dimension_results} reports Macro-F1 and ordinal
MAE for each quality dimension.
Macro-F1 measures classification performance across the low,
medium, and high levels.
Ordinal MAE measures the distance between predicted and reference
levels using the encoding $0$, $1$, and $2$.

\begin{table}[t]
    \centering
    \caption{
        Prediction of individual quality dimensions.
        P, G, M, S, and E denote task progress, goal correctness,
        manipulation quality, safety, and execution stability,
        respectively.
    }
    \label{tab:supp_dimension_results}
    \small
    \setlength{\tabcolsep}{3pt}
    \begin{tabular*}{\columnwidth}{
        @{\extracolsep{\fill}}lccccc@{}
    }
        \toprule
        Metric & P & G & M & S & E \\
        \midrule
        Macro-F1 $\uparrow$
        & .748 & .740 & .732 & .745 & .758 \\
        Ordinal MAE $\downarrow$
        & .248 & .247 & .222 & .216 & .208 \\
        \bottomrule
    \end{tabular*}
\end{table}

Macro-F1 ranges from $0.732$ to $0.758$, and ordinal MAE ranges
from $0.208$ to $0.248$.
Execution stability has the highest Macro-F1 and the lowest
ordinal MAE.
Manipulation quality has a lower Macro-F1 than task progress
and goal correctness but a lower ordinal MAE, reflecting the
different aspects of classification performance and ordinal
prediction error captured by these metrics.

\subsection{Reward Supervision Ablations}
\label{app:reward_ablations}

Table~\ref{tab:reward_evaluation} in the main paper compares
Scalar, 2D, and 5D supervision.
Scalar predicts utility alone.
2D adds supervision for task progress and goal correctness,
while 5D uses all five dimensions.
The three configurations share the same reference utility
and pairwise supervision, and rank candidates using a separate
utility prediction head.

We perform two independent ablations of the complete 5D model.
The first removes the pairwise loss while retaining dimension
prediction and utility regression.
The second shuffles the manipulation quality, safety, and
execution stability labels to examine the contribution of
their correspondence with the associated executions.

Removing the pairwise loss reduces Top-1 accuracy from
$33.07\%$ to $30.99\%$.
Shuffling the three dimension labels reduces Top-1 accuracy
to $29.17\%$ and mean selected execution utility from
$0.4603$ to $0.4550$.
Both relative ranking supervision within candidate pools
and the correspondence between quality labels and executions
contribute to candidate ranking.

\subsection{Reward Input Information}
\label{app:input_information}

We compare three input configurations:
the initial image, task instruction, and candidate actions;
the same inputs with an imagined video;
and the same inputs with a recorded execution video.
A reward model is trained separately for each configuration
and selects candidates from the same pools by predicted utility.
Table~\ref{tab:supp_input_information} reports the full results.
The imagined video row corresponds to the 5D model
in Table~\ref{tab:reward_evaluation} of the main paper.

\begin{table*}[t]
    \centering
    \caption{
        Candidate selection with different reward inputs.
        Pairwise and Top-1 accuracies are percentages.
        Utility is the mean execution reference utility of
        selected candidates across all 384 states.
        Below Source counts selections with lower execution
        reference utility than Source from the same initial state.
    }
    \label{tab:supp_input_information}
    \small
    \setlength{\tabcolsep}{5pt}
    \begin{tabular*}{\textwidth}{
        @{\extracolsep{\fill}}lrrrr@{}
    }
        \toprule
        Reward model input
        & Pairwise $\uparrow$
        & Top-1 $\uparrow$
        & Utility $\uparrow$
        & Below Source $\downarrow$ \\
        \midrule
        Initial image + instruction + actions
        & 55.91 & 29.95 & .4527 & 233 \\
        Same inputs + imagined video
        & 57.79 & 33.07 & .4603 & 219 \\
        Same inputs + recorded execution video
        & \textbf{64.32}
        & \textbf{50.52}
        & \textbf{.4735}
        & \textbf{165} \\
        \bottomrule
    \end{tabular*}
\end{table*}

Adding imagined video improves pairwise accuracy by
$1.88$ percentage points and Top-1 accuracy by
$3.12$ percentage points.
Mean selected execution utility increases from
$0.4527$ to $0.4603$.
These improvements indicate that imagined videos provide
useful information for candidate evaluation beyond the
initial image and action sequence.

Recorded execution videos yield the highest ranking accuracies
and mean selected execution utility.
Their Top-1 accuracy is $50.52\%$, exceeding the imagined video
configuration by $17.45$ percentage points.
This difference reflects a gap between evaluating predicted
futures and observing actual execution outcomes.
On the same candidate pools, retaining Source throughout
gives a mean utility of $0.4444$, while selecting the candidate
with the highest execution reference utility in each pool
gives $0.4871$.

\section{Additional Auditing and Selection Analysis}
\label{app:auditing}

This section supplements the risk identification and candidate
selection results in Sec.~\ref{sec:reliability_evaluation}
and examines the effect of different audit training targets.
The evaluation uses the 310 non-Source candidates selected
by reward ranking on the natural test set.
High-score mismatch and Below Source are defined
in Appendix~\ref{app:audit_targets}.

\subsection{Risk Identification and Group Analysis}
\label{app:risk_identification}

Table~\ref{tab:supp_audit_diagnostics}(a) reports the discrimination
performance of audit risk for the two events.
Among the 310 candidates, 23 satisfy the mismatch conditions,
and 219 have lower execution reference utility than their
corresponding Source.
Restricting predicted utility to at least $0.58$ yields
a high-score subset of 142 candidates containing all
23 mismatch cases.

\begin{table*}[t]
    \centering
    \caption{
        Event discrimination and grouping by audit risk.
        (a) The high-score subset contains candidates with
        predicted utility of at least $0.58$.
        (b) Candidates are ranked by audit risk.
        Percentages in parentheses are event rates within each group.
    }
    \label{tab:supp_audit_diagnostics}
    \small
    \setlength{\tabcolsep}{6pt}

    \begin{tabular*}{\textwidth}{
        @{\extracolsep{\fill}}lccc@{}
    }
        \multicolumn{4}{@{}l}{
            \textbf{(a) Event discrimination}
        } \\
        \toprule
        Event and population
        & Positives / Candidates
        & AUROC $\uparrow$
        & AP $\uparrow$ \\
        \midrule
        Below Source, all proposals
        & 219/310 & .774 & .872 \\
        High-score mismatch, all proposals
        & 23/310 & .77 & .30 \\
        High-score mismatch, high-score subset
        & 23/142 & .73 & .38 \\
        \bottomrule
    \end{tabular*}

    \par\medskip

    \begin{tabular*}{\textwidth}{
        @{\extracolsep{\fill}}lccc@{}
    }
        \multicolumn{4}{@{}l}{
            \textbf{(b) Groups ranked by audit risk}
        } \\
        \toprule
        Group
        & Candidates
        & High-score mismatch
        & Below Source \\
        \midrule
        All proposals
        & 310 & 23 (7.42\%) & 219 (70.65\%) \\
        Highest-risk 20\%
        & 62 & 14 (22.58\%) & 50 (80.65\%) \\
        Lowest-risk 20\%
        & 62 & 1 (1.61\%) & 26 (41.94\%) \\
        \bottomrule
    \end{tabular*}
\end{table*}

Within the high-score subset, the auditor achieves a mismatch
AUROC of $0.73$, distinguishing risk among candidates that
already receive high reward predictions.
Mismatch prevalence is $7.42\%$ among all proposals and
$16.20\%$ within the high-score subset.
We report positive counts alongside AP to specify the
evaluation population for each result.

The 62 candidates with the highest audit risk contain
14 mismatch cases, whereas the 62 with the lowest risk
contain one.
The Below Source rate also decreases from $80.65\%$
in the highest-risk group to $41.94\%$ in the lowest-risk group.
Audit risk therefore captures differences in both prediction
mismatch and replacement quality.

\subsection{Selection at a Fixed Replacement Count}
\label{app:fixed_replacements}

We select 111 candidates from the same 310 non-Source reward
winners and retain Source at the remaining initial states.
All rules use the same replacement count without an advantage
margin.
Mean execution utility is computed over all 384 test states,
including those that retain Source.

Table~\ref{tab:supp_fixed_replacements} extends the comparison
in Table~\ref{tab:reliability_evaluation}(b) of the main paper.
Random selection reports the expected outcome of uniformly
selecting 111 candidates.
The auditor without utility removes predicted utility and
its derived features while retaining the other inputs.

\begin{table}[t]
    \centering
    \caption{
        Selection with 111 replacements.
        Better and Worse count replacements with higher and
        lower execution reference utility than Source, respectively.
        Utility is averaged over all 384 test states.
    }
    \label{tab:supp_fixed_replacements}
    \small
    \setlength{\tabcolsep}{3pt}
    \begin{tabular*}{\columnwidth}{
        @{\extracolsep{\fill}}lrrr@{}
    }
        \toprule
        Selection rule
        & Better $\uparrow$
        & Worse $\downarrow$
        & Utility $\uparrow$ \\
        \midrule
        Random (expected)
        & 32.58 & 78.42 & .4501 \\
        Predicted advantage
        & 47 & 64 & .4589 \\
        Simple score risk
        & 46 & 65 & .4602 \\
        Logistic regression risk
        & 52 & 59 & .4621 \\
        Auditor w/o utility
        & 51 & 60 & .4621 \\
        Full auditor
        & \textbf{58} & \textbf{53} & \textbf{.4636} \\
        \bottomrule
    \end{tabular*}
\end{table}

The full auditor retains 11 more beneficial replacements
than predicted advantage ranking and six more than logistic
regression risk.
Without utility features, the auditor still retains
51 beneficial replacements, compared with 47 for predicted
advantage.
Including utility information increases this count to 58.

This comparison selects a fixed number of candidates by score
ranking.
Default TIF instead determines acceptance using the advantage
margin and risk threshold.
Both procedures produce 111 replacements on this test set,
but they select different candidate sets.

\subsection{Effect of the Audit Training Target}
\label{app:audit_target_analysis}

We compare signed-error regression, Below-Source classification,
and high-score mismatch classification.
The three models share auditor inputs, training data,
and training budget.
Their targets are defined in Appendix~\ref{app:audit_targets}.
Table~\ref{tab:supp_audit_targets} reports event discrimination
on the same 310 candidates.

\begin{table*}[t]
    \centering
    \caption{
        Event discrimination with different audit training targets.
    }
    \label{tab:supp_audit_targets}
    \small
    \setlength{\tabcolsep}{6pt}
    \begin{tabular*}{\textwidth}{
        @{\extracolsep{\fill}}lccc@{}
    }
        \toprule
        Training target
        & Mismatch AUROC $\uparrow$
        & Mismatch AP $\uparrow$
        & Below Source AUROC $\uparrow$ \\
        \midrule
        Signed-error regression
        & .73 & .25 & .75 \\
        Below-Source classification
        & .64 & .16 & \textbf{.79} \\
        High-score mismatch classification
        & \textbf{.77} & \textbf{.30} & .77 \\
        \bottomrule
    \end{tabular*}
\end{table*}

Below-Source classification achieves the highest AUROC
for its corresponding event, at $0.79$.
High-score mismatch classification achieves the highest
mismatch AUROC and AP, at $0.77$ and $0.30$, respectively.
Its Below Source AUROC in
Table~\ref{tab:supp_audit_targets} is the same result as
in Table~\ref{tab:supp_audit_diagnostics}(a),
displayed at a different precision.

Table~4(a) of the main paper reports the corresponding
selection quality and downstream policy performance.
When accepting 111 test candidates, Below-Source classification
retains the fewest Worse replacements, at 50.
High-score mismatch classification retains 53 but achieves
a higher mean execution utility of $0.4636$.
Worse counts utility decreases, whereas mean utility also
accounts for their magnitudes, so the two metrics can
rank selectors differently.

The downstream comparison uses natural training candidates,
with replacement counts matched by suite.
Each training set contains 230 replacement trajectories
and 570 Source trajectories.
Signed-error regression, Below-Source classification,
and high-score mismatch classification achieve mean policy
success rates of $75.60\%$, $75.85\%$, and $76.25\%$,
respectively.
In this comparison, the mismatch target achieves both
the highest mean execution utility in offline selection
and the highest downstream policy success rate.

\section{Detailed Policy Results}
\label{app:policy}

This section supplements the policy evaluation in the main paper.
All methods use the training configuration in
Appendix~\ref{app:selection_training}.
Each training seed is evaluated over 2,000 episodes across
the four LIBERO suites.
Table~\ref{tab:supp_seed_results} reports success counts
for individual seeds.
Means and sample standard deviations are computed from
the success rates of the three seeds.

\subsection{Results Across Seeds and Matched Replacement Counts}
\label{app:policy_seeds}

Policy comparisons at matched replacement counts use
the 348 natural training proposals that satisfy the advantage margin.
Each method accepts 57, 61, 54, and 58 replacements in
Spatial, Object, Goal, and Long, respectively,
while retaining 200 training trajectories per suite.
Each resulting training set contains 230 replacement trajectories
and 570 Source trajectories.

Predicted advantage ranking prioritizes candidates with larger
advantages, while logistic regression risk and audit risk
prioritize candidates with lower predicted risk.
Random-K samples the required number of candidates within each suite.
Low-entropy-K and Consistency-K select candidates according
to prediction entropy and consistency scores, respectively.
For each method, the selected set remains unchanged
across the three training seeds.

\begin{table*}[t]
    \centering
    \caption{
        Policy success counts for individual training seeds.
        Each seed is evaluated over 2,000 episodes.
        Mean success rates are reported in percent with
        sample standard deviations.
        The TIF row also supplies the results for matched
        replacement comparisons and the high-score mismatch target.
    }
    \label{tab:supp_seed_results}
    \small
    \setlength{\tabcolsep}{6pt}
    \begin{tabular*}{\textwidth}{
        @{\extracolsep{\fill}}lrrrr@{}
    }
        \toprule
        Configuration
        & Seed 7
        & Seed 11
        & Seed 23
        & Mean success (\%) \\
        \midrule

        \multicolumn{5}{@{}l}{
            \textit{Reference configurations and reward supervision}
        } \\
        Source
        & 1464 & 1479 & 1452 & $73.25 \pm 0.68$ \\
        2D complete pipeline
        & 1507 & 1519 & 1500 & $75.43 \pm 0.48$ \\
        TIF (5D)
        & 1521 & 1538 & 1516 & $76.25 \pm 0.58$ \\

        \midrule
        \multicolumn{5}{@{}l}{
            \textit{Selection rules at matched replacement counts}
        } \\
        Random-K
        & 1484 & 1495 & 1470 & $74.15 \pm 0.63$ \\
        Advantage Top-K
        & 1511 & 1521 & 1501 & $75.55 \pm 0.50$ \\
        Logistic risk-K
        & 1507 & 1525 & 1507 & $75.65 \pm 0.52$ \\
        Low-entropy-K
        & 1498 & 1512 & 1492 & $75.03 \pm 0.51$ \\
        Consistency-K
        & 1506 & 1524 & 1506 & $75.60 \pm 0.52$ \\

        \midrule
        \multicolumn{5}{@{}l}{
            \textit{Audit targets at matched replacement counts}
        } \\
        Signed-error regression
        & 1510 & 1522 & 1504 & $75.60 \pm 0.46$ \\
        Below-Source classification
        & 1514 & 1530 & 1507 & $75.85 \pm 0.59$ \\

        \bottomrule
    \end{tabular*}
\end{table*}

The TIF row uses the same evaluation records as the main
policy experiment.
Across all three seeds, TIF exceeds both predicted advantage
ranking and logistic regression risk selection.
Its gains over predicted advantage are $0.50$, $0.85$,
and $0.75$ percentage points, respectively.
The corresponding gains over logistic regression risk
are $0.70$, $0.65$, and $0.45$ percentage points.

The audit target comparison also matches replacement counts
within each suite.
Signed-error regression selects candidates in ascending order
of predicted error, while the two classification targets
select in ascending order of predicted risk.
This comparison matches counts through ranking;
the default TIF risk threshold of $0.50$ is not applied
to the regression scores.

\subsection{Effect of Reward Dimensions on the Long Suite}
\label{app:policy_long}

The main paper compares the average performance of Scalar,
2D, and 5D within the complete selection and policy training
pipeline.
Table~\ref{tab:supp_long_results} provides the Long suite results.
All configurations use the same reference utility, gating rules,
and policy training budget, with a separate auditor trained
for each reward model.

\begin{table}[t]
    \centering
    \caption{
        Policy success on Long and averaged across all four suites
        with different reward supervision.
        Values are percentages, reported as mean $\pm$ sample
        standard deviation across three training seeds.
    }
    \label{tab:supp_long_results}
    \small
    \setlength{\tabcolsep}{5pt}
    \begin{tabular*}{\columnwidth}{
        @{\extracolsep{\fill}}lcc@{}
    }
        \toprule
        Supervision & Long & Average \\
        \midrule
        Scalar
        & $43.33 \pm 0.92$
        & $74.57 \pm 0.33$ \\
        2D
        & $46.13 \pm 1.30$
        & $75.43 \pm 0.48$ \\
        5D
        & $\mathbf{48.73 \pm 0.83}$
        & $\mathbf{76.25 \pm 0.58}$ \\
        \bottomrule
    \end{tabular*}
\end{table}

On Long, 5D improves over Scalar and 2D by $5.40$ and
$2.60$ percentage points, respectively.
These gains exceed the corresponding improvements of
$1.68$ and $0.82$ percentage points averaged across all four suites.
The effect of dimensional supervision varies across suites.
For example, on Goal, 5D achieves $82.73\%$ mean success,
compared with $83.47\%$ for 2D.

\subsection{Supervision from Different Trajectory Steps}
\label{app:trajectory_steps}

The comparison between first-action supervision and supervision
from all valid steps uses the same retained trajectories
within each selection method.
Policy initialization, update count, and batch size
are also held fixed.
The first-action setting uses only the initial observation--action
pair of each segment.
The all-valid-steps setting samples a segment uniformly
and then samples a valid time step uniformly within it.

The complete results are reported in Table~4(c) of the main paper.
Using all valid steps increases mean success from
$71.97\%$ to $73.25\%$ for Source and from
$74.18\%$ to $76.25\%$ for TIF.
Under the same update budget, observations and actions
from later steps provide additional training information.

\section{Real-Robot Evaluation Details}
\label{app:robot}

This section supplements the real-robot candidate selection
experiment in the main paper.
We use a JAKA S5 manipulator to record candidates with
different execution quality on three tabletop tasks.
Recorded execution videos are used to evaluate reward ranking
and audit-based selection.

\subsection{Tasks and Candidate Pools}
\label{app:robot_tasks}

The three tasks are pushing a block across a target line,
placing exactly one cup, and relocating a specified cup.
Each task contains 20 candidate pools.
Each pool contains five physically executed trajectories,
one designated as Source, yielding 60 pools and
300 execution trajectories in total.

The pools cover different degrees of task progress,
completion quality, and execution deviations,
providing alternative behaviors for the same task.
Reward-only and TIF each select one trajectory per pool,
giving 60 selections per method.
Table~\ref{tab:supp_robot_tasks} defines the tasks.

\begin{table*}[t]
    \centering
    \caption{
        Real-robot tasks, completion criteria,
        and aspects of execution quality.
    }
    \label{tab:supp_robot_tasks}
    \small
    \setlength{\tabcolsep}{5pt}
    \begin{tabular}{@{}p{0.18\textwidth}
                      p{0.39\textwidth}
                      p{0.35\textwidth}@{}}
        \toprule
        Task
        & Goal and completion criteria
        & Execution quality \\
        \midrule

        Push across line
        & Push the green block completely beyond the blue
          target line and leave it in a stable state.
        & Whether the block fully crosses the line and whether
          execution collides with or displaces the obstacle.
        \\[4pt]

        Exactly-one-cup placement
        & Place exactly one cup upright and fully supported
          on the target platform beyond the blue line.
        & Compliance with the single-cup constraint,
          interaction with additional cups,
          and placement stability.
        \\[4pt]

        Specified-cup relocation
        & Move the specified cup from the green platform
          to the yellow platform and place it upright.
        & Selection of the specified cup, attempts to manipulate
          the wrong cup, and final placement stability.
        \\
        \bottomrule
    \end{tabular}
\end{table*}

In the push task, the blue outlined box marks the block's
initial position, while the blue target line defines
the crossing condition.
Completion of the cup placement task also depends on
compliance with the quantity constraint during execution.
Manipulating two cups violates the instruction,
even if the final image appears close to the target state.

\subsection{Video Evaluation and Candidate Selection}
\label{app:robot_selection}

The real-robot experiment uses recorded execution videos
as visual inputs.
Executions of different quality provide candidate outcomes
for comparing reward evaluation and audit selection
in the JAKA setting.

Both methods use the same reward model.
Reward-only selects the candidate with the highest
predicted utility.
TIF uses the same auditor and acceptance rule as in
the main paper.
It first fixes the candidate with the highest predicted utility,
then checks whether its predicted advantage over Source
exceeds $0.15$ and its audit risk is at most $0.50$.
The candidate replaces Source when both conditions hold;
otherwise, Source is retained.
Model parameters and selection thresholds remain unchanged
throughout the real-robot evaluation.

We inspect the corresponding execution videos and record
two outcomes:
\begin{itemize}
    \item \textbf{Task completion:}
    The execution satisfies the task goal and instruction
    constraints in Table~\ref{tab:supp_robot_tasks}.

    \item \textbf{Clean completion:}
    The task is completed with a stable final state and without
    collision, wrong- or extra-object interaction,
    or unnecessary recovery.
\end{itemize}

Clean completion is a subset of task completion.
An execution that reaches the goal after a collision may
count as task completion but not clean completion.
The evaluation unit is the execution selected by each method
from each candidate pool.

\subsection{Selection Outcomes and Example}
\label{app:robot_results}

Table~6 of the main paper summarizes the results across
the three tasks.
Among 60 selections, Reward-only and TIF select
45 and 47 trajectories that complete the task,
including 35 and 42 clean completions, respectively.
Task completion increases from $75.00\%$ to $78.33\%$,
and clean completion increases from $58.33\%$ to $70.00\%$.

TIF selects three more clean completions for pushing,
two more for single-cup placement, and two more for
specified-cup relocation.
For relocation, task completion remains at $16/20$,
while clean completion increases from $13/20$ to $15/20$,
reflecting differences in execution quality beyond
terminal task completion.

Figure~\ref{fig:real_robot} in the main paper shows
a candidate pool from the push task.
Table~\ref{tab:supp_robot_example} lists its scores.

\begin{table}[t]
    \centering
    \caption{
        Scores for the five candidates in the push example
        shown in Fig.~\ref{fig:real_robot}.
        Predicted advantage is relative to Source C1.
    }
    \label{tab:supp_robot_example}
    \small
    \setlength{\tabcolsep}{4pt}
    \begin{tabular*}{\columnwidth}{
        @{\extracolsep{\fill}}lccc@{}
    }
        \toprule
        Candidate
        & Pred.\ utility
        & Audit risk
        & Advantage \\
        \midrule
        C1: Source & .62 & .14 & $0.00$ \\
        C2         & .83 & .72 & $+0.21$ \\
        C3         & .72 & .23 & $+0.10$ \\
        C4         & .56 & .18 & $-0.06$ \\
        C5         & .66 & .31 & $+0.04$ \\
        \bottomrule
    \end{tabular*}
\end{table}

C2 has the highest predicted utility and an advantage
of $0.21$, exceeding the advantage margin.
Both Reward-only and the rule using only reward
and the advantage margin therefore select C2.
Its audit risk is $0.72$, exceeding the threshold of $0.50$,
so TIF retains Source C1.

During execution, both C1 and C2 push the block beyond
the target line.
C2 collides with and displaces the obstacle,
whereas C1 completes the push without collision
and leaves the block stable.
This example shows how the audit gate changes
the supervision acceptance decision after reward ranking.
\end{document}